\documentclass[conference]{IEEEtran}

\usepackage{cite}
\usepackage{amsmath, amssymb, amsfonts}
\usepackage{graphicx}
\usepackage{textcomp}
\usepackage{multirow}
\usepackage{array}
\usepackage{booktabs}
\usepackage{rotating}
\usepackage{hyperref}
\usepackage{fancyhdr}
\usepackage{lmodern}
\usepackage{adjustbox}
\usepackage{ragged2e}
\usepackage{balance}

\usepackage[table,xcdraw]{xcolor}
\usepackage{colortbl} 
\hypersetup{
    colorlinks=true,
    urlcolor=blue,
    citecolor=black,
    pdftitle={Overleaf Example},
    pdfpagemode=FullScreen,
}

\newcommand\blfootnote[1]{%
  \begingroup
  \renewcommand\thefootnote{}\footnote{#1}%
  \addtocounter{footnote}{-1}%
  \endgroup
}

\author{
	\IEEEauthorblockN{
		Palak Handa\IEEEauthorrefmark{1}, 
		Amirreza Mahbod\IEEEauthorrefmark{1}, 
		Florian Schwarzhans\IEEEauthorrefmark{1}, 
		Ramona Woitek\IEEEauthorrefmark{1}, 
		Nidhi Goel\IEEEauthorrefmark{2}, \\
		Manas Dhir\IEEEauthorrefmark{4}, 
		Deepti Chhabra\IEEEauthorrefmark{3}, 
		Shreshtha Jha\IEEEauthorrefmark{2}, 
		Pallavi Sharma\IEEEauthorrefmark{2},
		Vijay Thakur\IEEEauthorrefmark{5}, \\
        Simarpreet Singh Chawla \IEEEauthorrefmark{6}, 
		Deepak Gunjan\IEEEauthorrefmark{7}, 
		Jagadeesh Kakarla\IEEEauthorrefmark{8}, 
		Balasubramanian Raman\IEEEauthorrefmark{9}
	}
	\vspace{5mm}
	\IEEEauthorblockA{\IEEEauthorrefmark{1}Research Center for Medical Image Analysis and Artificial Intelligence, 
		Department of Medicine, \\Danube Private University, Krems, Austria}
	\IEEEauthorblockA{\IEEEauthorrefmark{2}Department of Electronics and Communication Engineering, 
		Indira Gandhi Delhi Technical \\University for Women, Delhi, India}
	\IEEEauthorblockA{\IEEEauthorrefmark{3}Department of Artificial Intelligence and Data Sciences, 
		Indira Gandhi Delhi Technical \\University for Women, Delhi, India}
	\IEEEauthorblockA{\IEEEauthorrefmark{4}Department of Artificial Intelligence and Machine Learning, 
		University School of Automation and \\Robotics, Guru Gobind Singh Indraprastha University, Delhi, India}
	\IEEEauthorblockA{\IEEEauthorrefmark{5}Department of Electronics and Communication Engineering, Delhi Technological University, Delhi, India}
    	\IEEEauthorblockA{\IEEEauthorrefmark{6}Columbia University, New York, NY, USA}
	\IEEEauthorblockA{\IEEEauthorrefmark{7}Department of Gastroenterology and HNU, 
		All India Institute of Medical Sciences, Delhi, India}
	\IEEEauthorblockA{\IEEEauthorrefmark{8}Department of Computer Science and Engineering, 
		Indian Institute of Information Technology,\\ Design and Manufacturing (IIITDM), Kancheepuram, Chennai, India}
	\IEEEauthorblockA{\IEEEauthorrefmark{9}Department of Computer Science and Engineering, 
		Indian Institute of Technology Roorkee, India}
}

\title{Capsule Vision 2024 Challenge: Multi-Class Abnormality Classification for Video Capsule Endoscopy}

\begin{document}

\maketitle

\begin{abstract}
We present the Capsule Vision 2024 Challenge: Multi-Class Abnormality Classification for Video Capsule Endoscopy. It was virtually organized by the Research Center for Medical Image Analysis and Artificial Intelligence (MIAAI), Department of Medicine, Danube Private University, Krems, Austria in collaboration with the 9th International Conference on Computer Vision \& Image Processing (CVIP 2024) being organized by the Indian Institute of Information Technology, Design and Manufacturing (IIITDM) Kancheepuram, Chennai, India. This document provides an overview of the challenge, including the registration process, rules, submission format, description of the datasets used, qualified team rankings, all team descriptions, and the benchmarking results reported by the organizers.
\end{abstract}


\blfootnote{Important key links of the challenge:}
\blfootnote{Challenge related questions: \href{mailto:ask.misahub@gmail.com}{ask.misahub@gmail.com}}
\blfootnote{Challenge hosting website: \href{https://misahub2023.github.io/cv2024.html}{Link}}
\blfootnote{Challenge GitHub repository: \href{https://github.com/misahub2023/Capsule-Vision-2024-Challenge}{Link}}
\blfootnote{Challenge ArXiv: \href{https://arxiv.org/abs/2408.04940}{Link}}
\blfootnote{Training and validation dataset: \href{https://figshare.com/articles/dataset/Training_and_Validation_Dataset_of_Capsule_Vision_2024_Challenge/26403469?file=48018562}{Link}}
\blfootnote{Test dataset: \href{https://figshare.com/articles/dataset/Testing_Dataset_of_Capsule_Vision_2024_Challenge/27200664?file=49717386}{Link}}
\blfootnote{Sanity Checker: \href{https://capsulevisionchallengesanitychecker.streamlit.app/}{Link}}
\blfootnote{Submission report Overleaf: \href{https://www.overleaf.com/read/kwhvpznnbzwb\#26d62a}{Link}}

\section{Overview of the Challenge}
A significant rise in the burden of Gastro-Intestinal (GI) and liver diseases across the globe has been observed due to varying environmental factors brought on by industrialization, changes in nutrition, and increased use of antibiotics \cite{arnold2020global, shah2018burden, yu2024differences}. Different endoscopic techniques are utilized in the diagnosis, treatment, and management of these diseases. Video Capsule Endoscopy (VCE) is one such technique that allows direct visualization of the GI tract and travels with the help of a disposable capsule-shaped device \cite{iddan2000wireless}. The device comprises an optical dome, a battery, an illuminator, an imaging sensor, and a transmitter.

VCE is a non-invasive technique that does not suffer from any sedation-related complications and has improved a physician’s ability to find different anomalies in the GI tract, especially small bowel-related diseases such as Crohn’s disease, Celiac disease, and intestinal cancer, among others \cite{iddan2000wireless, thwaites2024current}. It is often considered as an alternative to conventional endoscopy techniques \cite{thwaites2024current, rondonotti2005complications}.

The true potential of VCE technology is hindered by the longer reading time of the video frames without compromising the quality of the report and cost of the capsule \cite{gupta2018journey}. During 6-8 hours of VCE procedure, a video of the GI tract trajectory is recorded on a device attached to the patient’s belt, which produces about 57,000-1,000,000 frames \cite{goel2022dilated}. These video frames are analysed retrospectively by experienced gastroenterologists. Presently, an experienced gastroenterologist takes approximately 2–3 hours to inspect the captured VCE video through a frame-by-frame analysis \cite{goel2022dilated}. This analysis is subject to human bias and high false-positive rates due to bubbles, debris, intestinal fluid, foreign objects, and chyme (food), among other factors negatively affecting the mucosal frames \cite{lazaridis2021implementation}. Further, the inadequate doctor-to-patient ratio across the globe delays this process. Other hardware-related technological limitations include capsule retention, battery limitations, and bowel obstructions \cite{rondonotti2005complications, gupta2018journey, goel2022dilated}.

Artificial intelligence (AI) is predicted to have profound effects on the future of VCE technology in the context of abnormality classification \cite{messmann2022expected}. There is a need for investigation and state-of-the-art development of robust, user-friendly, interpretable AI-based models for multi-class abnormality classification that can aid in reducing the burden on gastroenterologists and save their valuable time by reducing the inspection time of VCE frames while maintaining high diagnostic precision.

\subsection{Significance of the Challenge}
\begin{itemize}
	\item The aim of the challenge was to provide an opportunity for the development, testing, and evaluation of AI models for automatic classification of abnormalities captured in VCE video frames.
	\item This challenge consisted of distinct training, validation, and test datasets for training, internal validation, and external validation purposes. 
	\item It promoted the development of vendor-independent and generalized AI-based models for an automatic abnormality classification pipeline with 10 class labels, namely: angioectasia, bleeding, erosion, erythema, foreign body, lymphangiectasia, polyp, ulcer, worms, and normal.
\end{itemize}

\subsection{Prizes}
\begin{itemize}
	\item Cash prize sponsored by the 9\textsuperscript{th} International Conference on Computer Vision \& Image Processing (CVIP 2024).
	\begin{itemize}
		\item 1\textsuperscript{st} Prize: 200 \texteuro
		\item 2\textsuperscript{nd} Prize: 150 \texteuro
		\item 3\textsuperscript{rd} Prize: 100 \texteuro
	\end{itemize}
	\item E-certificate to all the ranked teams.
	\item Co-authorship in the challenge summary paper for the top 4 teams.
	\item Opportunity to showcase work at CVIP 2024.
\end{itemize}

\subsection{Challenge relevant dates}
\begin{itemize}
	\item Launch of the challenge: August 15, 2024
	\item Registration open: August 15, 2024 - October 10, 2024
	\item Release of Training Data: August 15, 2024
	\item Release of Test Data and Result submission open: October 11, 2024 - October 25, 2024 
	\item Result analysis by the organizing team: October 26, 2024 - November 24, 2024
	\item Announcement of results for all teams: November 25, 2024
\end{itemize}

\section{Registration and Rules}
\subsection{Rules for participation}
\begin{itemize}
	\item This challenge was open to all students (B. Tech/M.Tech/Ph.D. of all branches), faculty members, researchers, clinicians, and industry professionals across the globe for free. 
	\item It took place in full-virtual mode.
	\item Participants were either registered as a solo participant or could form a team.
	\item The results were asked to be submitted in a specific format over email to \href{mailto:ask.misahub@gmail.com}{ask.misahub@gmail.com}. The template for submission was provided.

\end{itemize}

\subsection{Rules for Team Formation}
\begin{itemize}
	\item A team could have a maximum of 4 participants.
	\item Team members could be from the same or different organizations or affiliations.
	\item A participant could only be a part of a single team.
	\item Only one member of the team was allowed to register for the challenge.
	\item Each team could only have a single registration. Multiple registrations per team led to disqualification.
	\item There were no limitations on the number of teams from the same organizations or affiliations. (However, each participant could only be part of a single team.)
\end{itemize}

\subsection{Rules for use of Training and Validation Dataset}
\begin{itemize}
	\item Download the dataset. It consists of the labelled training and validation dataset. 
	\item Develop a model to classify 10 class labels using ONLY the training and validation dataset. 
	\item Store the model, associated weights, and files. Compute the predictions and evaluation metrics on the validation dataset (internal validation) as per the provided sample code on our github repository. 
	\item The participants were allowed to perform augmentations, resize the data for saving computational resources, and could use existing machine learning, meta-learning, ensembles, any pre-trained models, and deep learning models. 
	\item A sample code was provided on our github how to load the dataset. 
\end{itemize}

\subsection{Rules for use of Testing Dataset}
\begin{itemize}
	\item Download the dataset. It consisted of the unlabeled test dataset.
	\item  Compute the predictions (external validation) as per the provided sample code on our github repository.
	\item Store the developed Excel sheet and submit it as a part of your submission.
	\item Prepare the report and github repository as per the submission format. 
\end{itemize}

\subsection{Submission format}
Each team was required to submit their results in an email with the following structure to \href{mailto:ask.misahub@gmail.com}{ask.misahub@gmail.com}.
\begin{itemize}
	\item The email should contain:
	\begin{itemize}
		\item Challenge name and team name as the SUBJECT LINE.
		\item Team member names and affiliations in the BODY OF THE EMAIL.
		\item Contact number and email address in the BODY OF THE EMAIL.
		\item A link to the github repository in public mode in the BODY OF THE EMAIL.
		\item A link to their report on any open preprint server of their choice (ArXiv, Authorea, BioRxiv, Figshare, etc.)  in the BODY OF THE EMAIL.
		\item Generated Excel sheet of the predicted train, validation, and test dataset (in xlsx format) as an attachment.
	\end{itemize}
	\item The github repository in public mode should contain the following:
	\begin{itemize}
		\item Developed code for training, validation, and testing in .py or .mat, etc. in readable format with comments.
		\item Stored model, associated weights or files \\(optional).
		\item Any utils, assets, or config. or checkpoints.
	\end{itemize}
\end{itemize}

\subsection{Important Notes}
\begin{itemize}
	\item The challenge dates and the time zones were considered to be as per the Central European Summer Time (CEST; UTC+02:00, Austria). They were counted as per end of the day 23:59. 
	\item The github repository should be public. Repositories that require access were NOT considered for evaluation.
	\item The submitted repository MUST be readable, documented properly, and interactive.
	\item The participants were requested to STRICTLY follow the submission format.
	\item The participants were NOT allowed to utilize any other dataset for training, validation, or testing in this challenge. 
	\item The participants were strictly advised to use the overleaf template for developing their report. The zip file is also available at our github repository.  
	\item The participants were requested to acknowledge and cite the datasets in their reports and for any other research purposes. Use of the datasets for commercial products or any purpose other than research is STRICTLY NOT ALLOWED.
	\item All participating teams have allowed the organizing team to summarize their developed AI pipelines in the form of a challenge summary paper.
	\item Any submitted Excel sheet with missing data was evaluated but given zero marks.
	\item For the final test set result submission, both training and validation datasets were allowed for training.
	\item For the final test set result submission, only one submission was allowed. In the case of multiple submissions, only the last one was considered.
	\item The organizing team had conducted a preliminary analysis utilizing a variety of machine learning models. The sample codes were made available on our github repository. 
	\item All teams were asked to verify their generated excel sheets using the sanity checker. 
\end{itemize}

\begin{figure*}
	\centering
	\includegraphics[width=\linewidth]{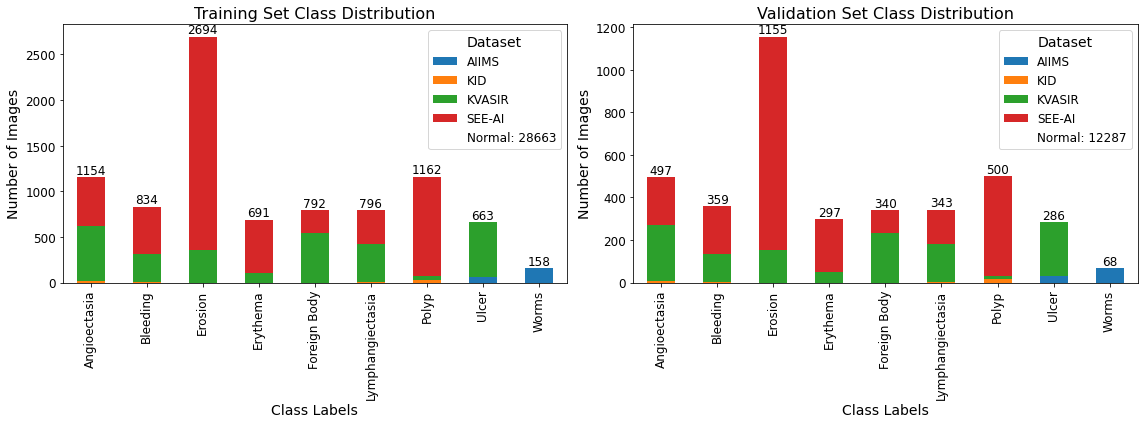}
	\caption{Class-wise distribution of the training and validation datasets. The classes represent all the abnormalities, including angioectasia, bleeding, erosion, erythema, foreign body, lymphangiectasia, polyps, ulcers, and worms.}
	\label{fig:data_vis_1}
	
\end{figure*}

\begin{figure*}
	\centering
	\includegraphics[width=0.9\linewidth]{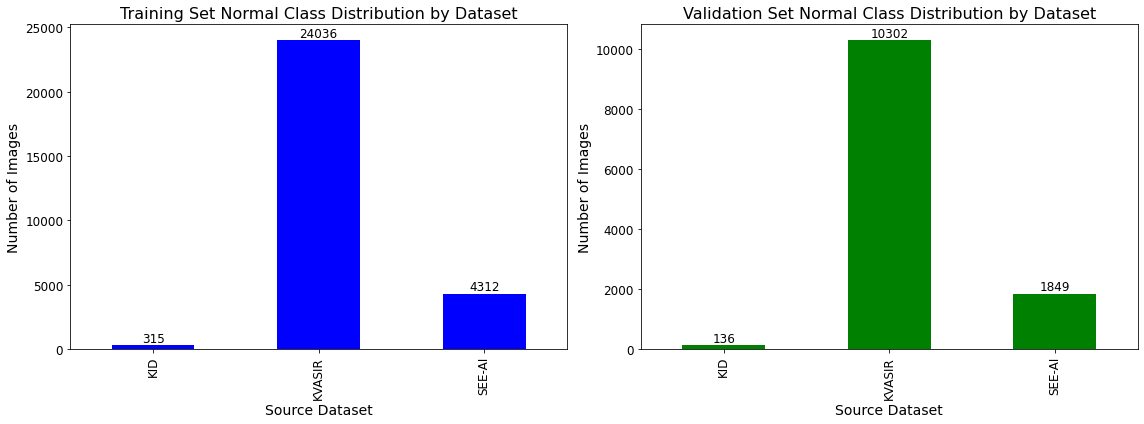}
	\caption{Class-wise distribution of the normal VCE frames in the training and validation datasets.}
	\label{fig:data_vis_2}
\end{figure*}

\begin{figure*}
	\centering
	\includegraphics[width=0.6\linewidth]{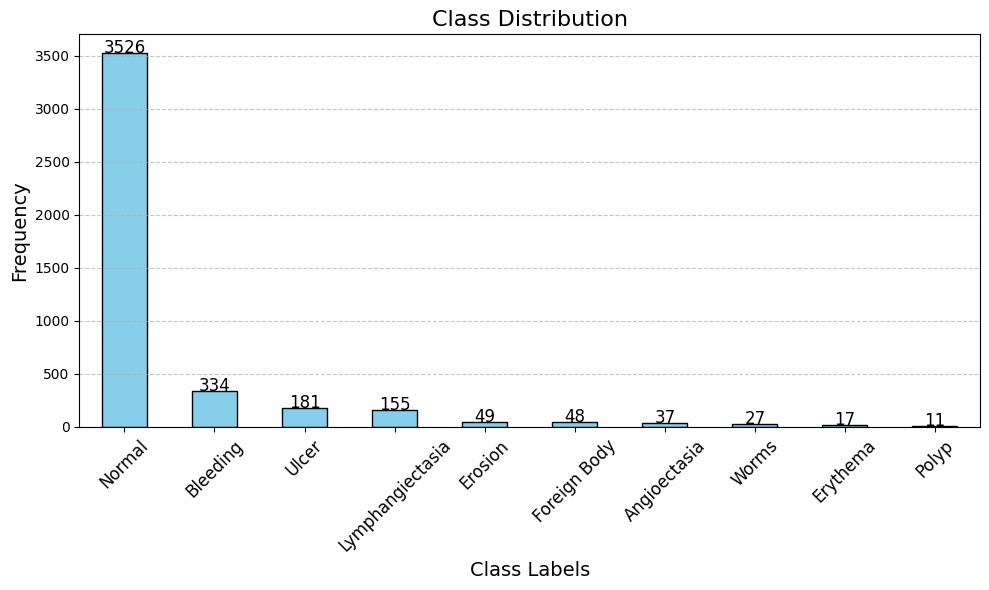}
	\caption{Class-wise distribution of the testing dataset.}
	\label{fig:test}
\end{figure*}

\begin{figure*}
	\centering
	\includegraphics[width=0.9\linewidth]{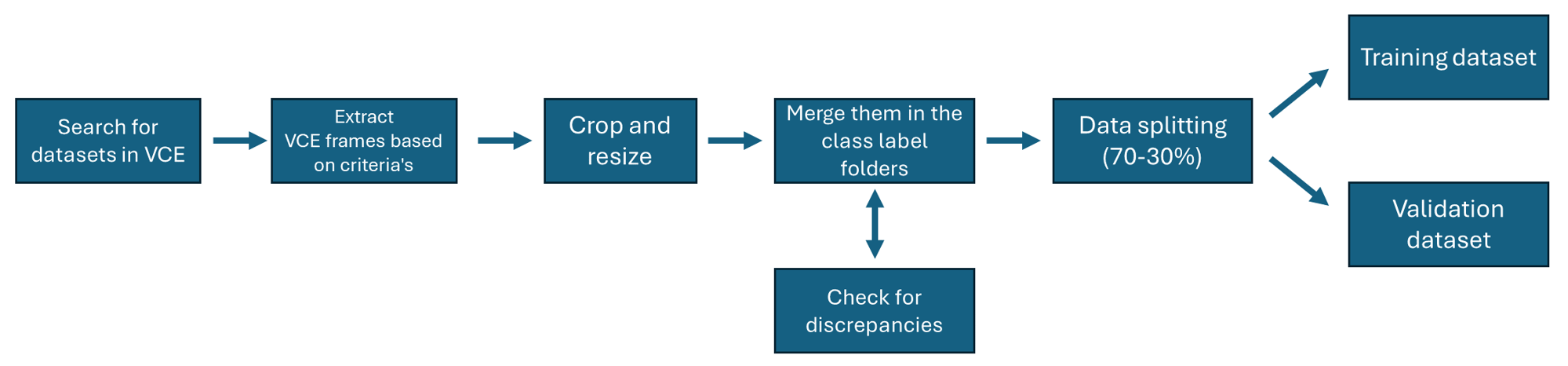}
	\caption{Pipeline for developing the training and validation datasets for this challenge. A similar pipeline was also utilized for development of the testing dataset.}
	\label{fig:pipeline}
\end{figure*}

\begin{table*}[htbp]
	\caption{No. of VCE frames with respect to their class labels and source dataset in the training, validation, and test datasets.}
	\label{data_details}
	\resizebox{\textwidth}{!}{%
		\begin{tabular}{llllllllllll}
			\hline
			Type of Data     & Source Dataset & Angioectasia & Bleeding & Erosion & Erythema & Foreign Body & Lymphangiectasia & Normal & Polyp & Ulcer & Worms \\ 
			\hline
			& KID     & 18           & 3        & 0       & 0        & 0            & 6                & 315    & 34    & 0     & 0     \\
			& KVASIR  & 606          & 312      & 354     & 111      & 543          & 414              & 24036  & 38    & 597   & 0     \\
			Training \cite{Handa2024training} & SEE-AI  & 530          & 519      & 2340    & 580      & 249          & 376              & 4312   & 1090  & 0     & 0     \\
			& AIIMS   & 0            & 0        & 0       & 0        & 0            & 0                & 0      & 0     & 66    & 158   \\ \hline
			Total Frames     &         & 1154         & 834      & 2694    & 691      & 792          & 796              & 28663  & 1162  & 663   & 158   \\
			\hline
			& KID     & 9            & 2        & 0       & 0        & 0            & 3                & 136    & 15    & 0     & 0     \\
			& KVASIR  & 260          & 134      & 152     & 48       & 233          & 178              & 10302  & 17    & 257   & 0     \\
			Validation \cite{Handa2024training}  & SEE-AI  & 228          & 223      & 1003    & 249      & 107          & 162              & 1849   & 468   & 0     & 0     \\
			& AIIMS   & 0            & 0        & 0       & 0        & 0            & 0                & 0      & 0     & 29    & 68    \\ \hline
			Total Frames     &         & 497          & 359      & 1155    & 297      & 340          & 343              & 12287  & 500   & 286   & 68  \\
			\hline 
			
			Testing \cite{Handa2024testing} & AIIMS    &         37  &     334    &   49     &     17 &      48
			&  155 &  3526 &  11   &   181   & 27     \\ \hline
	\end{tabular}}
\end{table*}

\subsection{Criteria of judging a submission}
\begin{itemize}
	\item All emails received before the result submission closing date were considered for evaluation.
	\item The github repository, report on any open-source pre-print server, and excel sheets received for all entries\\ were downloaded by the organizing team.
	\item A common data file was prepared to compare the evaluation metrics achieved for training, validation, testing the dataset. It is now available on our github.
	\item A semi-automated Python script file was used to achieve the best evaluation metrics received among all entries and is available on our github.
	\item The following checklist was used to select the top three winning teams:
	\begin{itemize}
		\item Combined metric on testing dataset. The combined metric was calculated by averaging the mean AUC and balanced accuracy for each team, separately for each dataset. It yielded a value between 0 and 1, with higher values indicating better performance, reflecting a balance between how well the model distinguishes each class (mean AUC) and the accuracy across classes (balanced accuracy).
	\end{itemize}
\end{itemize}

\section{Datasets}

\subsection{Training, validation, and test dataset description}
The training and validation dataset were developed using three publicly available (SEE-AI project dataset \cite{yokote2024small}, KID \cite{koulaouzidis2017kid}, and Kvasir-Capsule dataset \cite{smedsrud2021kvasir}) and one private (AIIMS \cite{goel2022dilated}) VCE dataset. The training and validation datasets consisted of 37,607 and 16,132 VCE frames, respectively, mapped to 10 class labels, namely angioectasia, bleeding, erosion, erythema, foreign body, lymphangiectasia, polyp, ulcer, worms, and normal. Fig.~\ref{fig:data_vis_1} and Fig.~\ref{fig:data_vis_2} depict the class distribution of the training and validation as per the source datasets. The source datasets in the training and validation datasets were ONLY extracted, cropped, resized, and split for the development of a large dataset in this challenge.

The testing dataset was developed using a high-quality, medically annotated private dataset from the Department of Gastroenterology and HNU, All India Institute of Medical Sciences, Delhi, India. It consisted of 4,385 VCE frames, which have been collected from more than 70 patient VCE videos with abnormalities, namely angioectasia, bleeding, erosion, erythema, foreign body, lymphangiectasia, polyp, ulcer, and worms (see Fig. \ref{fig:test}). The images were de-identified and does not contain any patient information. The data collection for this AI study was done according to Helsinki declarations. It was approved by the Department of Gastroenterology and HNU, All India Institute of Medical Sciences Delhi, India ethics committee (Ref. No.: IEC-666/05.08.2022). Further, it is part of the objectives of the Core Research Grant Project (CRG/2022/001755). The ground truth labels of the test dataset are now released and updated on Figshare.

\subsection{Development of the Training and Validation Dataset}

A standard pipeline was followed to develop the training and validation dataset for this challenge (see Fig.~\ref{fig:pipeline}). At first, all the publicly available datasets in VCE research were searched and collected. The datasets found were kvasir-capsule \cite{smedsrud2021kvasir}, SEE-AI project \cite{yokote2024small}, EndoSLAM \cite{ozyoruk2021endoslam}, annotated bleeding dataset \cite{deeba2016automated}, rhodes island \cite{charoen2022rhode}, WCEBleedGen \cite{hub2024auto}, AutoWCEBleedGen-Test~\cite{hub2024auto}, AI-KODA~\cite{handa2024comprehensive}, KID \cite{koulaouzidis2017kid}, and red lesion endoscopy \cite{leenhardt2020cad}.

EndoSLAM, AI-KODA, and rhode island datasets were not considered as they do not contain abnormalities and focus on virtual capsule reconstruction, cleanliness assessment, and organ classification, respectively. Annotated bleeding dataset and red lesion endoscopy dataset were not considered as they only contain red lesions. WCEBleedGen was not considered as it contains frames from KID, kvasir-capsule, annotated bleeding dataset and only focuses on bleeding classification, detection, and segmentation. AutoWCEBleedGen-Test has been developed by the co-authors of this challenge and hence was considered as a part of the test dataset. All the frames were anonymized and renamed.

Finally, the KID, kvasir-capsule, and SEE-AI projects were considered. Then the VCE frames and their abnormalities in each of these datasets were downloaded and extracted based on the inclusion and exclusion criteria.

\begin{itemize}
	\item Inclusion criteria:
	\begin{itemize}
		\item VCE frames with one abnormality in each frame.
		\item Normal VCE frames.
	\end{itemize}
	\item Exclusion criteria:
	\begin{itemize}
		\item VCE frames with multiple abnormalities and duplicates.
	\end{itemize}
\end{itemize}

\begin{table*}[htbp]
	\caption{Results sorted in descending order based on the combined metric calculated for the testing set by the organizing team.}
	\centering
	\label{tab1}
	% \resizebox{\textwidth}{!}{%
		\begin{tabular}{lllll}
			\hline
			\textbf{Rank} & \textbf{Team Name}  & \textbf{Mean AUC}  & \textbf{Balanced Accuracy}  & \textbf{Combined Metric} \\
			\hline
			1   & PuppyOps \cite{PuppyOps}     & 0.8570334 & 0.3573283 & 0.6071809 \\
			2   & MedInfoLab IIT Hyderabad \cite{MedInfoLab} & 0.7736101 & 0.3719341  & 0.5727721 \\
			3   & WueVision \cite{WueVision}    & 0.7625353 & 0.3710301 & 0.5667827   \\
			4   & Llama\_Mamba \cite{Llama_Mamba} & 0.7632455 & 0.3366313  & 0.5499384   \\
			5   & Seq2Cure \cite{Seq2Cure}     & 0.7460799 & 0.3467811  & 0.5464305   \\
			6   & Taaldhwaj \cite{Taaldhwaj}    & 0.7271482 & 0.3359341 & 0.5315412   \\
			7   & Capsule Commandos \cite{Capsule_Commandos}  & 0.7313885  & 0.3234582 & 0.5274234 \\
			8   & eAI \cite{eAI}          & 0.7486922 & 0.2759186 & 0.5123054 \\
			9   & VCap \cite{VCap}         & 0.7364247 & 0.2813258 & 0.5088752 \\
			10  & DS \& Chill   & 0.8197379 & 0.1918587  & 0.5057983 \\
			11  & Layer Players \cite{Layer_Players} & 0.6585702 & 0.3509284 & 0.5047493 \\
			12  & Rookies \cite{Rookies}      & 0.7237849 & 0.2617322 & 0.4927586 \\
			13  & Team\_CSIR \cite{Team_CSIR}   & 0.6968434 & 0.2465738 & 0.4717086 \\
			14  & CapsuleNet \cite{CapsuleNet}   & 0.6724393 & 0.2674035 & 0.4699214 \\
			15  & Code Cortex \cite{Code_Cortex}  & 0.6821873 & 0.2298121 & 0.4559997 \\
			16  & ViFo Tech \cite{ViFo_Tech}    & 0.6166294 & 0.2826887 & 0.4496591 \\
			17  & 1b2w \cite{1b2w}         & 0.6426787  & 0.2363980 & 0.4395384 \\
			18  & Machine Minds \cite{Machine_Minds} & 0.6980735 & 0.1786562  & 0.4383648 \\
			19  & aiVengers     & 0.6432135  & 0.2270782 & 0.4351459 \\
			20  & Optiminds     & 0.7113022 & 0.1541079 & 0.4327051 \\
			21  & Pioneers \cite{Pioneers}     & 0.7065916 & 0.1195959  & 0.4130937 \\
			22  & Organic       & 0.6011660 & 0.2231773  & 0.4121716 \\
			23  & STEM sisters \cite{STEM_sisters}  & 0.5821777 & 0.1398035 & 0.3609906 \\
			24  & DeepScope Innovators \cite{DeepScope}  & 0.5632873   & 0.1083343 & 0.3358108 \\
			25  & BotBotBot \cite{BotBotBot}     & 0.5212472 & 0.1354561 & 0.3283516 \\
			26  & Deep\_Learners \cite{Deap_Learners} & 0.5118418 & 0.1194708  & 0.3156563 \\
			27  & EndoAI  \cite{EndoAI}       & 0.4974579 & 0.0766949 & 0.2870764 \\
			\hline
		\end{tabular}
		% }
\end{table*}

\begin{table*}[htbp]
	\caption{Benchmarking results calculated by the organizing team on the test dataset.}
	\label{tab3}
	\centering
	% \resizebox{\textwidth}{!}{%
		\begin{tabular}{ccccc}
			\hline
			\textbf{S. No.} & \textbf{Model Name}  & \textbf{Mean AUC}  & \textbf{Balanced Accuracy}  & \textbf{Combined Metric} \\
			\hline
			1 &VGG19 &0.5254758 &0.1444602 & 0.3349680\\
			2 &Xception &0.5341090 &0.1313365 & 0.3327228\\
			3 &ResNet50V2 &0.5421599 &0.1772507 & 0.3597053\\
			4 &MobileNetV2 &0.5484505 &0.1140326 &0.3312415 \\
			5 &InceptionV3 &0.5249649 &0.1284446 &0.3267047 \\
			6 &InceptionResNetV2 &0.5232011 &0.1469184 &0.3350598
			\\
			\hline
		\end{tabular}
		% }
\end{table*}

A new class label of worms with 226 VCE frames was introduced in the training and validation dataset. VCE frames containing 95 ulcer VCE frames were also added. The ulcer and worm VCE frames were collected from three different patients, respectively, from the department of gastroenterology, HNU, All India Institute of Medical Sciences, Delhi, India, and were also utilized in the study conducted by Goel et al. \cite{goel2022dilated}.

All the frames were cropped and resized to $224\times224$ pixel resolution. The frames were merged as per their source and class label folders and checked for discrepancies. Checks for discrepancies included that all VCE frames should be:

\begin{itemize}
	\item Correctly labelled as per their class name and in the right source folder.
	\item Of $224\times224$ resolution and in .jpg format.
	\item Properly cropped from the four-side black boundaries.
	\item NOT have any patient information or annotation markings.
	\item NOT have multiple abnormalities. 
	\item NOT have duplicates.
\end{itemize}

After a thorough check, the main folder was subjected to a random data split of 70-30\% for the development of the training and validation dataset. Table~\ref{data_details} details the number of VCE frames with respect to their respective class labels in each of the datasets. Similar standard pipeline was followed for the development of the test dataset.

\section{Results}

This challenge received about 150 team participations across the globe during the registration stage, out of which 35 teams finally submitted their results. After resolving several discrepancies in initial submissions and requesting corrected files, the organizing team finalized a total of 27 qualified teams for evaluation. Table \ref{tab1} details the rankings calculated by the organizing team of 27 qualified entries based on their performance on the test dataset. Table \ref{tab3} details the benchmarking results calculated by the organizing team on the test dataset. No data augmentation, nor fine-tuning of the the AI models was performed in the benchmarking experiments. We are providing a description of all the 35 received entries in the Annexure A and also release the results for the train, validation, and test datasets for all teams, and our benchmark experiments through Table \ref{tab:BrTable} and \ref{tab:BenchmarkingTable}. All supporting codes, received excel sheets, reports of all teams and result evaluation done by the organizing team is available on our github.

\section{Conclusion}

The Capsule Vision 2024 Challenge seeked to advance VCE through a multi-class abnormality classification. By offering unique training, validation, and test datasets, the challenge promoted the development of efficient models to aid gastroenterologists in reducing analysis time while maintaining high diagnostic accuracy. We invited participants to use these resources, strictly follow submission guidelines, and contribute to enhancing VCE abnormality classification. This challenge acted as a platform for innovation and collaboration in medical imaging, aiming for impactful advancements of AI in VCE. 

\section{Acknowledgment}

We appreciate the prize money by CVIP 2024 and support by the Research Center for MIAAI, Department of Medicine, Danube Private University, Krems, Austria, MISAHUB, IIITDM Kancheepuram, Chennai, India, AIIMS New Delhi, India, and DST SERB, India for helping to conduct this challenge.

\section*{Annexure A}

\subsection{Description of all the received entries}

\textbf{Team PuppyOps} utilized DINOv2, combining a vision transformer for feature extraction and fully connected layers for multi-class classification. Random flips and brightness adjustments were applied as data augmentation techniques. \href{https://github.com/Sri-Karthik-Avala/Capsule-Challenge-2024}{Link to access code}.

\textbf{Team MedInfoLab IIT Hyderabad} performed fine-tuning of BiomedCLIP-PubMedBERT by integrating the PubMedBERT language model with a ViT, which is followed by BiomedCLIP generating visual and textual inputs for better classification. They also performed random transformations to generate augmented data for training. \href{https://github.com/Satyajithchary/MedInfoLab_Capsule_Vision_2024_Challenge}{Link to access code}.  

\textbf{Team WueVision} applied domain-adaptive pre-training on the EVA-02 mode, adapting it for the general task of GIE using their custom dataset, EndoExtend24 - a fusion of ten public and private endoscopy datasets. To address class imbalance, they employed data augmentations and weighted sampling. \href{https://github.com/mvrcii/capsule_vision_challenge_2024}{Link to access code}. 

\textbf{Team Llama\_Mamba} implemented FasterViT-3, an optimized ViT model known for computational efficiency. They adapted the model by modifying the final classification layer to align with the Capsule dataset's categories and used pre-trained weights for initialization. No explicit class imbalance mitigation techniques or unique methodologies were mentioned. \href{https://github.com/manoj060603/Capsule_Vision_2024}{Link to access code}.

\textbf{Team Seq2Cure} proposed a multi-model ensemble combining CNN and transformer architectures, leveraging complementary strengths through full parameter fine-tuning for all models. They addressed class imbalance with weighted random sampling, focal loss, and data augmentations. \href{https://github.com/arnavs04/capsule-vision-2024}{Link to access code}.

\textbf{Team Taaldhwaj} proposed a CAVE-Net (Classifying Abnormalities in Video Capsule Endoscopy Network), an ensemble image classification model. It combines predictions from three models using soft voting: a CBAM-enhanced ResNet, a DNN based on a pre-trained ResNet-50 autoencoder, and Syn-XRF, which includes SVM, Random Forest, KNN, and XGBoost. Data augmentation techniques like flips, rotations, zooming, cropping, shearing, elastic transformations, Gaussian noise, and image sharpening were used to balance the dataset and create up to 7,500 samples per class. \href{https://github.com/NextGenAI-Lab/taaldhwaj-vision-challenge-2024}{Link to access code}.

\textbf{Team Capsule Commandos} concluded with DaViT (Dual Attention Vision Transformers) with two attention blocks—spatial window attention and channel group attention, as their chosen model for the challenge while also implementing custom CNN, ResNet-50, ViT, and Multiscale ViT on the training and validation dataset. \href{https://github.com/devrishivermaa/capsule-commandos}{Link to access code}.

\textbf{Team eAI} implemented EfficientViT-l2, an optimized ViT model known for computational efficiency. The team used pre-trained weights for initialization and mitigated class imbalance with weighted sampling, weighted cross-entropy loss, and augmentations such as MixUp. \href{https://github.com/GirinChutia/Misahub-Capsule-Vision-2024-Solution-eAI}{Link to access code}. 

\textbf{Team VCap} proposed a multi-backbone feature extraction approach incorporating ResNet50, DeiT, and MobileNetV3-Large. They employed SMOTE for addressing class imbalance in under-represented classes. Their use of diverse backbone architectures and targeted re-sampling techniques highlighted a unique and balanced strategy. \href{https://github.com/reubenjrouse/VCap_CapsuleVision2024}{Link to access code}.

\textbf{Team DS\&Chill} implemented a hybrid ensemble model combining YOLOv11 for object detection and DenseNet121 for classification, leveraging a soft voting mechanism to integrate predictions from both models. They addressed class imbalance by augmenting the dataset by merging the training and validation dataset and re-splitting it along with targeted transformations. \href{https://github.com/cipherdoge/Misahub_CV24}{Link to access code}.

\textbf{Team Layer Players} incorporated Squeeze and Excitation blocks into the ResNet-101 architecture for training using cross-entropy loss and Adam optimizer with learning rate scheduler for optimized performance. They have performed standard preprocessing techniques of normalization and random transformations. \href{https://github.com/flux661/Capsule-Vision-2024-Solution-Team_Layer_Players}{Link to access code}.

\textbf{Team Rookies} implemented a Swin Transformer pre-trained on ImageNet for classifying gastrointestinal abnormalities in VCE images. While the report did not explicitly mention specific class imbalance mitigation techniques, the use of a pre-trained Swin Transformer inherently helped the model leverage learned features from large, diverse datasets like ImageNet, which can improve generalization and reduce overfitting. Furthermore, the team employed data augmentation techniques, such as random horizontal flips and image normalization, to increase the diversity of the training data, which could have helped mitigate the effects of any class imbalance by enhancing the model's robustness across different gastrointestinal conditions. These techniques, alongside the pre-trained model, contributed to the model’s overall performance despite the potential challenges posed by class imbalance. \href{https://github.com/Manav1115/Capsule-Vision-2024-Challenge}{Link to access code}.

\textbf{Team\_CSIR} explored multiple models, including ResNet-18, ResNet-50, DenseNet-121, DenseNet-169, and MobileNetV3, and concluded MobileNetV3 with a self-attention mechanism as it showed the best performance. The issue of class imbalance was addressed by implementing focal loss. \href{ https://github.com/ajay-pratap-singh-518/Capsule-Vision-2024-challenge}{Link to access code}. 

\textbf{Team CapsuleNet} fine-tuned EfficientNet-b7 with PReLU activation functions for better feature representation. Selective class-based augmentations ensured approximately 1500 samples per class to mitigate the imbalance. The use of PReLU activation introduced a unique optimization in learning non-linear features. \href{https://github.com/ayushman72/CapsuleNet}{Link to access code}.

\begin{table}[htbp]
	\centering
	\caption{All received entries results calculated by the organizing team on the train, validation, and test dataset.}
	\begin{sideways}  
		\label{tab:BrTable}
		\resizebox{2.4\columnwidth}{!}{  % Resize the table based on text height
			\begin{tabular}{ccccccccccccccccc}
				\hline
				\multirow{2}{*}{\textbf{S. No.}} & \multirow{2}{*}{\textbf{Team names}} 
				& \multicolumn{5}{c}{{\color[HTML]{1F1F1F} \textbf{Evaluation metrics generated for training dataset}}} 
				& \multicolumn{5}{c}{\textbf{Evaluation metrics generated for validation dataset}} 
				& \multicolumn{5}{c}{\textbf{Evaluation metrics generated for testing dataset}} \\
				\cline{3-17} 
				
				& &  \multicolumn{2}{c}{\textbf{Goal metrics}} & \multicolumn{3}{c}{\textbf{Other metrics}} & \multicolumn{2}{c}{\textbf{Goal metrics}} & \multicolumn{3}{c}{\textbf{Other metrics}} & \multicolumn{2}{c}{\textbf{Goal metrics}} & \multicolumn{3}{c}{\textbf{Other metrics}} \\
				\cmidrule(lr){3-4} \cmidrule(lr){5-7} \cmidrule(lr){8-9} \cmidrule(lr){10-12} \cmidrule(lr){13-14} \cmidrule(lr){15-17}  % Add cmidrule for each section
				
				& & \textbf{mean\_auc} & \textbf{balanced\_accuracy} & \textbf{avg\_precision} & \textbf{avg\_f1} & \textbf{avg\_specificity} 
				& \textbf{mean\_auc} & \textbf{balanced\_accuracy} & \textbf{avg\_precision} & \textbf{avg\_f1} & \textbf{avg\_specificity} 
				& \textbf{mean\_auc} & \textbf{balanced\_accuracy} & \textbf{avg\_precision} & \textbf{avg\_f1} & \textbf{avg\_specificity} \\
				\hline
				
				1                                & \textbf{eAI} \cite{eAI}                          & 0.997441                & 0.949012                         & 0.866819                     & 0.904149             & 0.993671                      & 0.982055           & 0.826765                    & 0.756554                & 0.78875          & 0.986985                  & 0.748692           & 0.275919                    & 0.21471                 & 0.144262         & 0.904503                  \\
				2                                & \textbf{Code Cortex} \cite{Code_Cortex}                  & 0.999717                & 0.961211                         & 0.965866                     & 0.963172             & 0.997572                      & 0.983253           & 0.777146                    & 0.802235                & 0.787479         & 0.983279                  & 0.682187           & 0.229812                    & 0.207601                & 0.160119         & 0.910863                  \\
				3                                & \textbf{EndoVisionaries}              & \multicolumn{15}{c}{Train and validation excel sheets NOT submitted.}                                                                                                                                                                                                                                                                                                                                          \\
				4                                & \textbf{WueVision} \cite{WueVision}                    & 0.997079                & 0.983182                         & 0.950182                     & 0.965683             & 0.998585                      & 0.991844           & 0.915906                    & 0.897701                & 0.906106         & 0.993692                  & 0.762535           & 0.37103                     & 0.276021                & 0.240386         & 0.95479                   \\
				5                                & \textbf{Seq2Cure} \cite{Seq2Cure}                     & 0.999723                & 0.982743                         & 0.952345                     & 0.96693              & 0.998306                      & 0.989574           & 0.863437                    & 0.866642                & 0.864494         & 0.989961                  & 0.74608            & 0.346781                    & 0.274025                & 0.237353         & 0.947611                  \\
				6        & \textbf{CapsuleNet} \cite{CapsuleNet}                   & 0.95317                 & 0.755215                         & 0.473999                     & 0.488614             & 0.951719                      & 0.945947           & 0.742682                    & 0.458138                & 0.47376          & 0.950721                  & 0.672439           & 0.267404                    & 0.187029                & 0.118891         & 0.908286                  \\
				7                                & \textbf{DeepScope Innovators} \cite{DeepScope}         & 0.623443                & 0.136505                         & 0.130489                     & 0.071668             & 0.928075                      & 0.613658           & 0.137861                    & 0.099039                & 0.072467         & 0.928091                  & 0.563287           & 0.108334                    & 0.099931                & 0.027602         & 0.906523                  \\
				8                                & \textbf{EndoAI} \cite{EndoAI}                       & 0.501353                & 0.101739                         & 0.10155                      & 0.030809             & 0.899906                      & 0.508361           & 0.104361                    & 0.098942                & 0.048532         & 0.900062                  & 0.497458           & 0.076695                    & 0.096228                & 0.01665          & 0.899755                  \\
				9                                & \textbf{Optiminds}                     & 0.963262                & 0.700381                         & 0.723253                     & 0.697683             & 0.974212                      & 0.974431           & 0.764223                    & 0.773863                & 0.756628         & 0.976929                  & 0.711302           & 0.154108                    & 0.267043                & 0.062794         & 0.864557                  \\
				10                               & \textbf{Pioneers} \cite{Pioneers}                     & \multicolumn{10}{c}{Submitted train and validation excel sheet format is NOT correct.}                                                                                                                                                                                             & 0.706592           & 0.119596                    & 0.26328                 & 0.152408         & 0.919396                  \\
				11                               & \textbf{Rookies} \cite{Rookies}                      & 0.504658                & 0.098811                         & 0.098669                     & 0.098709             & 0.900434                      & 0.992521           & 0.831267                    & 0.845258                & 0.835782         & 0.988487                  & 0.723785           & 0.261732                    & 0.24331                 & 0.155366         & 0.886288                  \\
				12                               & \textbf{Machine Minds} \cite{Machine_Minds}                & \multicolumn{10}{c}{Submitted train and validation excel sheet format is NOT correct.}                                                                                                                                                                                             & 0.698074           & 0.178656                    & 0.229997                & 0.104646         & 0.902981                  \\
				13                               & \textbf{BIOMIL}                       & \multicolumn{15}{c}{Excel sheets NOT submitted.}                                                                                                                                                                                                                                                                                                                                                         \\
				14                               & \textbf{Fixcells}                     & \multicolumn{15}{c}{Excel sheets NOT submitted.}                                                                                                                                                                                                                                                                                                                                                         \\
				15                               & \textbf{ViFo Tech} \cite{ViFo_Tech}                    & 0.491587                & 0.1                              & 0.003069                     & 0.005954             & 0.9                           & 0.480729           & 0.1                         & 0.003081                & 0.005978         & 0.9                       & 0.616629           & 0.282689                    & 0.098016                & 0.10433          & 0.901583                  \\
				16                               & \textbf{VCap} \cite{VCap}                         & 0.992147                & 0.852361                         & 0.845333                     & 0.840494             & 0.986844                      & 0.986745           & 0.830575                    & 0.797365                & 0.806828         & 0.987198                  & 0.736425           & 0.281326                    & 0.274523                & 0.214663         & 0.947931                  \\
				17                               & \textbf{Llama\_Mamba} \cite{Llama_Mamba}                 & 0.999812                & 0.970652                         & 0.970135                     & 0.970374             & 0.998772                      & 0.993089           & 0.869323                    & 0.87398                 & 0.871459         & 0.991077                  & 0.763246           & 0.336631                    & 0.274797                & 0.24259          & 0.956782                  \\
				18                               & \textbf{BotBotBot} \cite{BotBotBot}                    & 0.48755                 & 0.10251                          & 0.120676                     & 0.091468             & 0.901431                      & 0.466174           & 0.102746                    & 0.128895                & 0.091898         & 0.901315                  & 0.521247           & 0.135456                    & 0.179776                & 0.052777         & 0.895732                  \\
				19                               & \textbf{BioTriads}                    & \multicolumn{15}{c}{Excel sheets NOT submitted.}                                                                                                                                                                                                                                                                                                                                                         \\
				20                               & \textbf{STEM sisters} \cite{STEM_sisters}                 & 1                       & 1                                & 1                            & 1                    & 1                             & 1                  & 1                           & 1                       & 1                & 1                         & 0.582178           & 0.139804                    & 0.11421                 & 0.048916         & 0.902093                  \\
				21                               & \textbf{MedInfoLab IIT Hyderabad} \cite{MedInfoLab}     & 0.999025                & 0.938807                         & 0.939549                     & 0.938876             & 0.996537                      & 0.993971           & 0.846436                    & 0.862021                & 0.853919         & 0.988462                  & 0.77361            & 0.371934                    & 0.240103                & 0.218057         & 0.931994                  \\
				22                               & \textbf{LAYER PLAYERS} \cite{Layer_Players}                & 0.996803                & 0.91983                          & 0.841935                     & 0.870067             & 0.99236                       & 0.984378           & 0.779634                    & 0.757434                & 0.752829         & 0.986512                  & 0.65857            & 0.350928                    & 0.2331                  & 0.189279         & 0.927541                  \\
				23                               & \textbf{Team\_CSIR} \cite{Team_CSIR}                   & 0.999935                & 0.973715                         & 0.978388                     & 0.975658             & 0.999249                      & 0.991755           & 0.831241                    & 0.857884                & 0.843214         & 0.988321                  & 0.696843           & 0.246574                    & 0.252623                & 0.190546         & 0.919277                  \\
				24                               & \textbf{Organic}                       & 0.996724                & 0.869612                         & 0.908544                     & 0.885051             & 0.988204                      & 0.9646             & 0.609549                    & 0.698828                & 0.641333         & 0.96713                   & 0.601166           & 0.223177                    & 0.195818                & 0.15074          & 0.922919                  \\
				25                               & \textbf{PuppyOps} \cite{PuppyOps}                     & 0.994421                & 0.826001                         & 0.876151                     & 0.848718             & 0.988432                      & 0.991462           & 0.784999                    & 0.851629                & 0.814887         & 0.98471                   & 0.857033           & 0.357328                    & 0.303032                & 0.249544         & 0.956976                  \\
				26                               & \textbf{Deep\_Learners} \cite{Deap_Learners}               & 0.948599                & 0.308641                         & 0.44231                      & 0.290985             & 0.984284                      & 0.646828           & 0.1002                      & 0.17617                 & 0.086873         & 0.900026                  & 0.511842           & 0.119471                    & 0.192495                & 0.124511         & 0.908299                  \\
				27                               & \textbf{Taaldhwaj} \cite{Taaldhwaj}                    & 0.950595                & 0.910828                         & 0.692859                     & 0.772933             & 0.990362                      & 0.944392           & 0.899312                        & 0.672232                 & 0.754322         & 0.989472                   & 0.727148           & 0.335934                    & 0.27234                 & 0.249095         & 0.943116                  \\
				28                               & \textbf{Capsule Commandos} \cite{Capsule_Commandos}            & 0.999883                & 0.967762                         & 0.974453                     & 0.97092              & 0.998739                      & 0.993245           & 0.859512                    & 0.879141                & 0.868847         & 0.990352                  & 0.731389           & 0.323458                    & 0.253447                & 0.227417         & 0.936104                  \\
				29                               & \textbf{aiVengers}                     & 0.501667                & 0.100375                         & 0.100377                     & 0.100375             & 0.900275                      & 0.999791           & 0.964595                    & 0.963503                & 0.963958         & 0.998517                  & 0.643214           & 0.227078                    & 0.198621                & 0.129267         & 0.908982                  \\
				30                               & \textbf{DS \& Chill}                                          & 0.955078                & 0.789926                         & 0.458631                     & 0.515254             & 0.945457                      & 0.956415           & 0.80422                     & 0.463798                & 0.523492         & 0.945694                  & 0.819738           & 0.191859                    & 0.165768                & 0.055007         & 0.881185                  \\
				31                               & \textbf{CODE MORPHERS} \cite{CODE_MORPHERS}                                        & \multicolumn{15}{c}{Excel sheets NOT submitted.}                                                                                                                                                                                                                                                                                                                                                         \\
				32                               & \textbf{PHEONIX ARIZONA} \cite{PHEONIX_ARIZONA}                                      & \multicolumn{15}{c}{Excel sheets NOT submitted.}                                                                                                                                                                                                                                                                                                                                                         \\
				33                               & \textbf{1b2w} \cite{1b2w}                                                 & 0.997562                & 0.944882                         & 0.836502                     & 0.878891             & 0.992244                      & 0.988577           & 0.841901                    & 0.761046                & 0.790384         & 0.98671                   & 0.642679           & 0.236398                    & 0.164349                & 0.142054         & 0.922782                  \\
				34                               & \textbf{InnoV8rs} \cite{InnoV8rs}                                             & \multicolumn{15}{c}{Submitted excels are of WRONG format.}                                                                                                                                                                                                                                                                                                                                               \\
				35                               & \textbf{Pixel Surgeons}               & \multicolumn{15}{c}{Submitted excels are of WRONG format.}   \\\hline 
			\end{tabular}
		}
	\end{sideways}
\end{table}

\begin{table}[htbp]
	\centering
	\caption{Benchmarking results calculated by the organizing team on the train, validation, and test dataset.}
	\label{tab:BenchmarkingTable}
	\begin{sideways}  % Rotate the table 90 degrees
		\resizebox{2.4\columnwidth}{!}{  % Use \textheight for resizing in the portrait page layout
			\begin{tabular}{ccccccccccccccccc}
				
				\hline
				\multirow{5}{4em}{\textbf{S. No.}}&\multirow{5}{4em}{\textbf{Model Name}}  & \multicolumn{5}{c}{\textbf{Evaluation Metrics Generated for Training Dataset by Organizing Team}} 
				& \multicolumn{5}{c}{\textbf{Evaluation Metrics Generated for Validation Dataset by Organizing Team}} 
				& \multicolumn{5}{c}{\textbf{Evaluation Metrics Generated for Testing Dataset by Organizing Team}}  \\
				\cmidrule(lr){3-7} \cmidrule(lr){8-12} \cmidrule(lr){13-17}
				
				& & \multicolumn{2}{c}{\textbf{Goal Metrics}} & \multicolumn{3}{c}{\textbf{Other Metrics}} & \multicolumn{2}{c}{\textbf{Goal Metrics}} & \multicolumn{3}{c}{\textbf{Other Metrics}} & \multicolumn{2}{c}{\textbf{Goal Metrics}} & \multicolumn{3}{c}{\textbf{Other Metrics}} \\
				
				\cmidrule(lr){3-4} \cmidrule(lr){5-7} \cmidrule(lr){8-9} \cmidrule(lr){10-12} \cmidrule(lr){13-14} \cmidrule(lr){15-17} 
				& & \textbf{mean\_auc} & \textbf{balanced\_} & \textbf{avg\_precision} & \textbf{avg\_f1\_score} & \textbf{avg\_specificity} & 
				\textbf{mean\_auc} & \textbf{balanced\_} & \textbf{avg\_precision} & \textbf{avg\_f1\_score} & \textbf{avg\_specificity} &
				\textbf{mean\_auc} & \textbf{balanced\_} & \textbf{avg\_precision} & \textbf{avg\_f1\_score} & \textbf{avg\_specificity}
				\\
				& & &\textbf{accuracy} &  & &  & 
				&  \textbf{accuracy} &  & &  &
				&  \textbf{accuracy} & & &   \\
				\cmidrule(lr){1-1} \cmidrule(lr){2-2} \cmidrule(lr){3-3} \cmidrule(lr){4-4} \cmidrule(lr){5-5} \cmidrule(lr){6-6} \cmidrule(lr){7-7} \cmidrule(lr){8-8}
				\cmidrule(lr){9-9} \cmidrule(lr){10-10} \cmidrule(lr){11-11} \cmidrule(lr){12-12} \cmidrule(lr){13-13} \cmidrule(lr){14-14} \cmidrule(lr){15-15} \cmidrule(lr){16-16} \cmidrule(lr){17-17}
				1& VGG19 & 0.9963138 &0.9818597 &0.9741388 &0.9778581 &0.9992781
				& 0.8555799 & 0.6241962 & 0.6574182 & 0.6380015 & 0.9673447 
				& 0.5254758 & 0.1444602 & 0.1733453 & 0.0674384 & 0.8970598  \\
				
				2& Xception &0.9920766 &0.9663486 &0.9761869 &0.9708231 &0.9985544 
				&0.7505018 &0.4871864 &0.6196647 &0.5381061 &0.9551189
				&0.5341090 &0.1313365 &0.1182914 &0.1023228 &0.9086853 \\
				
				3& ResNet50V2 &0.9884365 &0.9732594 &0.9768656 &0.9747702 &0.9988956 
				&0.7971128 &0.6090536 &0.6667782 &0.6283159 &0.9664619
				&0.5421599 &0.1772507 &0.2025479 &0.0933522 &0.9089577 \\
				
				4& MobileNetV2 &0.9997063 &0.9719589 &0.9798020 &0.9756844 &0.9989355 
				&0.9257260 &0.5358164 &0.6552588 &0.5730695 &0.9593709
				&0.5484505 &0.1140326 &0.1397954 &0.0613603 &0.8705717 \\
				
				5& InceptionV3 &0.9932378 &0.9659432 &0.9053089 &0.9287167 &0.9955336 
				&0.7724807 &0.5087122 &0.5343706 &0.4994157 &0.9622115
				&0.5249649 &0.1284446 &0.1055226 &0.0819467 &0.9006027 \\
				
				6& InceptionResNetV2 &0.6796624 &0.3681719 &0.7560009 &0.3447172 &0.9468554 
				&0.6138730 &0.2530873 &0.4418623 &0.2299317 &0.9389513
				&0.5232011 &0.1469184 &0.0899191 &0.0599153 &0.8974349 \\ \hline
			\end{tabular}
		}
	\end{sideways}
\end{table}

\textbf{Team Code Cortex} developed a CASCRNet (Capsule Endoscopy-ASPP-SCR-Network), a hybrid model combining atrous spatial pyramid pooling blocks with shared channel residual blocks. The team used focal loss to mitigate class imbalance. The use of pooling blocks for efficient feature extraction added depth to their methodology. \href{https://github.com/Manvith-Prabhu/CASCRNet}{Link to access code}.

\textbf{Team ViFo Tech} implemented a hybrid ResNet-50 and SVM model. They handled class imbalance using data augmentation techniques, including rotations, flips, zooms, and brightness adjustments. Their integration of deep learning with traditional SVM classification offered a distinct perspective. \href{https://github.com/Sankari17/vifotech_capsule_vision_challenge}{Link to access code}.

\textbf{Team 1b2w} implemented transfer learning with Swin Transformer and applied the learning rate scheduler to adjust parameters and prevent overfitting. \href{https://github.com/kanishk-8/Misahub_vce/tree/main}{Link to access code}.

\textbf{Team Machine Minds} proposed an Efficient Fusion U-Net that integrated EfficientNet B7 with a U-Net encoder. They addressed class imbalance through Gaussian noise removal. \href{https://github.com/kancharlavamshi/Capsule-Vision}{Link to access code}.

\textbf{Team aiVengers} employed a Swin Tranformer model combined with self-supervised contrastive learning using a two-phase training strategy of contrastive pre-training followed by supervised fine-tuning, enhancing feature learning from limited labeled data. Class imbalance is mitigated through contrastive learning involving triplet-based training and data augmentation to improve learning for minority classes. \href{https://github.com/Priyanshu-5257/cvc-24-submission}{Link to access code}. 

\textbf{Team Optiminds} utilized multiple ViT models, including DeiT, DinoV2, and Swin Transformers, to fine-tune performance. No specific class imbalance techniques were reported. \href{https://github.com/optiminds00/capsule-endoscopy-submission}{Link to access code}. 

\textbf{Team Pioneers} utilized a pre-trained ViT model from HuggingFace, fine-tuned on ImageNet-21k. They employed a balanced DataLoader with weighted sampling to address class imbalance. Despite their efforts, their submission did not meet qualification standards. \href{https://github.com/Roshon02/Pioneers-vision-capsule}{Link to access code}.

\textbf{Team Organic} implemented a custom CNN architecture for multi-class classification of VCE frames. This model integrated multiple convolutional layers with varying kernel sizes ($3\times3$, $5\times5$, $7\times7$), max pooling, and strategic dropout application to enhance feature extraction and generalization for anomaly detection. The model was compiled using an SGD optimizer and categorical cross-entropy loss function, achieving low loss scores that indicated no overfitting. \href{https://github.com/RajKrishnaChy/capsule_vision}{Link to access code}.

\textbf{Team Stem Sisters} used a semi-supervised approach with transductive learning and GLCM-based texture features for multi-class VCE image classification, outperforming other methods like Random Forest and SVM. They addressed class imbalance through data augmentation for minority classes and under-sampling for overrepresented classes. Their unique use of transductive learning, which focuses on specific un-labelled data, proved effective. \href{https://github.com/DhivyaPradeep-2025/capsule_vision}{Link to access code}.

\textbf{Team DeepScope Innovators} employed an ensemble of DenseNet and ResNet for robust feature extraction. The team utilized normalization and augmentation techniques, such as random horizontal flips and rotations, to mitigate class imbalance. Their focus on ensemble methods added robustness to the classification task. \href{https://github.com/amansagar0403/Capsule-Vision-2024-Challenge}{Link to access code}.

\textbf{Team BotBotBot} implemented a custom ViT-CNN hybrid architecture, combining a ViT branch and a ResNet34 CNN branch, followed by a classification head that performs binary classification first, and then specific abnormality is classified. They applied three levels of augmentations (Heavy, Medium, and Light) based on the intensity of transformations. \href{https://github.com/typos12onlr/VCE_Chal_MisaHub2024}{Link to access code}. 

\textbf{Team Deep\_Learners} implemented FuseCaps, a Fusion-based Framework for Capsule Endoscopy. This hybrid feature extraction method combines CNN, specifically DenseNet, for feature extraction concatenated with MLP, which performed dimensionality reduction on the features extracted using Radiomics on the training dataset. This addressed the class imbalance of the dataset by capturing both deep and handmade representations. \href{https://github.com/i-am-tobi/Capsule_Vision_Challenge_2024}{Link to access code}.

\textbf{Team EndoAI} introduced a ResNet-18-based encoder-decoder framework augmented with omni dimensional gated attention and wavelet transformation modules. These additions aimed to enhance feature localization and extraction. Their integration of wavelet transformations was a unique aspect of their methodology. \href{https://github.com/09Srinivas2005/Capsule-Endoscopy-Multi-classification-via-Gated-Attention-and-Wavelet-Transformations}{Link to access code}.

\bibliographystyle{IEEEtran.bst}
\bibliography{main_references.bib} %

\end{document}